# Will it rain tomorrow?

Bilal Ahmed - 561539
Department of Computing and Information Systems,
The University of Melbourne, Victoria, Australia
bahmad@student.unimelb.edu.au

**Abstract**

With the availability of high precision digital sensors and cheap storage medium, it is not uncommon to find large amounts of data collected on almost all measurable attributes, both in nature and man-made habitats. Weather in particular has been an area of keen interest for researchers to develop more accurate and reliable prediction models. This paper presents a set of experiments which involve the use of prevalent machine learning techniques to build models to predict the day of the week given the weather data for that particular day i.e. temperature, wind, rain etc., and test their reliability across four cities in Australia {Brisbane, Adelaide, Perth, Hobart}. The results provide a comparison of accuracy of these machine learning techniques and their reliability to predict the day of the week by analysing the weather data. We then apply the models to predict weather conditions based on the available data.

1. **Introduction**

Weather is perhaps the most commonly encountered natural phenomenon which affects a large proportion of the human population on a daily basis. Given the large number of variables which may contribute to the overall weather of a given location, it is quite challenging to accurately predict what the weather would be like on a given day and the day of the week based on the given weather conditions.

For our experiments we train our classifiers using historical data to:

1. Predict the day of the week {Mon, Tue, Wed, Thu, Fri, Sat, Sun} by analysing the given weather conditions for that day which includes temperature, rain, wind and time of the year among other attributes.

2. Predict weather conditions for a given day i.e. the likelihood of rain, wind and temperature range.

3. Test the robustness of these models by applying them across various cities in Australia and compare their results.

2. **Classifiers**

The following classification algorithms have been used to build prediction models to perform the experiments:

- **Naïve Bayes (NB)** classifier is a simple probabilistic classifier based on applying Bayes' theorem with strong (naive) independence assumptions .i.e. the classifier assumes that the presence (or absence) of a particular feature of a class is unrelated to the presence (or absence) of any other feature, given the class variable. It is simple to build and fast to make decisions. It efficiently accommodates new data by changing the associated probabilities.

- **Random Forests (RF)** classifier is a variant of the decision tree classification model. It operates by constructing a multitude of decision trees at training time and outputting the class that is the mode of the classes output by individual trees. This method is similar to bagging in many respects but the construction of each tree is different to the standard decision tree method. Random Forests are shown to be one of the best classification methods experimentally.

- **J48** classifier is a variant of the decision tree classification model and is based on C4.5 algorithm. The C4.5 algorithm generates a classification-decision tree for the given data-set by recursive partitioning of data. The decision is grown using depth-first strategy. J48 employs two pruning methods to reduce the size of the generated decision trees. The first is known as *sub-tree replacement* and the second is termed *sub-tree raising*.

- **IB1** classifier is an instance based learner, based on simple Euclidean distance. IB1 uses a simple distance measure to find the training instance closest to the given test instance, and predicts the same class as the training instance. If multiple instances are the same (smallest) distance to the test instance, the first one found is used.

3. **Methodology**

The classifiers described in Section 2 are trained on a range of datasets to predict the day of the week based on the weather conditions. The algorithms are compared based on the accuracy of their results.

We further investigate the correlation between the discretisation techniques and the accuracy of the results.

### 3.1. Pre-processing

Following steps have been applied to pre-process the data-sets.

#### 3.1.1. Missing Values

The missing values for attributes in the dataset are replaced with the modes and means based on existing data. The *ReplaceMissingValues*[1] filter in Weka is used to replace values for missing attributes in the dataset. Adding the missing values provides a more complete dataset for the classifiers to be trained on.

#### 3.1.2. Discretisation

The following two techniques were applied to discretise the attributes which were originally in continuous form.

1. **Unsupervised Discretisation** is used to discretise attributes into the following "groups" or bins:
    - 10 bins – High resolution
    - 4 bins – Medium resolution
    - 2 bins – Low resolution
    - 1 bin (similar to supervised discretisation)

2. **Supervised Discretisation:** The classifiers are also trained on data discretised using supervised discretisation technique.

For instance following results are obtained when

---

[1] weka.filters.unsupervised.attribute.ReplaceMissingValues

---

**unsupervised discretisation** is applied to the attribute (F4) which represents the aggregate precipitation (in mm), given in the training data set for Brisbane.

- Rainfall data discretised into **10** separate categories or ranges, i.e. from 0 to 0.5 mm, 0.5 to 5.5 and so on.

| Name: F4 | | Type: Nominal |
|---|---|---|
| Missing: 0 (0%) | Distinct: 10 | Unique: 0 (0%) |
| No. | Label | Count |
| 1 | '(-inf-0.5]' | 11378 |
| 2 | '(0.5-5.5]' | 735 |
| 3 | '(5.5-12.5]' | 682 |
| 4 | '(12.5-29.5]' | 645 |
| 5 | '(29.5-59.5]' | 650 |
| 6 | '(59.5-115]' | 681 |
| 7 | '(115-184.389639]' | 372 |
| 8 | '(184.389639-184.8…' | 1170 |
| 9 | '(184.889639-452]' | 585 |
| 10 | '(452-inf)' | 582 |

- Rainfall data discretised into **4** separate categories or ranges.

| Name: F4 | | Type: Nominal |
|---|---|---|
| Missing: 0 (0%) | Distinct: 4 | Unique: 0 (0%) |
| No. | Label | Count |
| 1 | '(-inf-0.5]' | 11378 |
| 2 | '(0.5-26.5]' | 2039 |
| 3 | '(26.5-184.389639]' | 1726 |
| 4 | '(184.389639-inf)' | 2337 |

- Rainfall data discretised into **2** separate categories or range

| Name: F4 | | Type: Nominal |
|---|---|---|
| Missing: 0 (0%) | Distinct: 2 | Unique: 0 (0%) |
| No. | Label | Count |
| 1 | '(-inf-0.5]' | 11378 |
| 2 | '(0.5-inf)' | 6102 |

As can be observed higher bin values provide higher resolution in terms of categorisation. For example by discretising data into 10 bins we get a much **higher resolution** as compared to when the attribute values are discretised into 2 bins. In the latter case, the data is divided into two very broad categories .i.e. (0 - 0.5mm) and (0.5mm – higher) and hence provides results of **coarse resolution** for the attribute. For instance, using the 2 bins approach we can only predict how likely it was to rain either more or less than 0.5 mm, since we only have two categories (0 – 0.5mm) and (0.5 mm – higher). On the other hand discretising into 10 bins provides higher resolution results; which does not necessarily mean higher accuracy. We use this knowledge when we try to predict rain, temperature and wind for a given day as part of our experiments. This is further discussed in Section 4.2 and the results in Section 4.2 (Result Set 2) further elaborate on this discussion.

The choice of discretisation resolution depends on the task (context) and on the type of data used.

4. **Results**

**4.1 Result Set 1** – Predicting the day of the week {Mon … Sun}, {Weekday, Sat, Sun} and {Weekday, Weekend} using training and development data

The following section outlines the results of the experiments.

1. Predicting the day of the week {Mon, Tue, Wed, Thu, Fri, Sat, Sun} by analysing the given weather conditions.

    **Result:** Discretising the Year attribute (F1) into 2 Bins coupled with Random Forest classifier yielded the highest accuracy, at **16.01 % for Brisbane data**. The second and third best performing algorithms have been marked in the following table (Table 1):

| Brisbane | | | | |
|---|---|---|---|---|
| Discretisation | Naïve Bayes Simple | Random Forests | J48 | IB1 |
| F1 to F20 – Supervised discretised | 14.53 % | 14.53 % | 14.53% | **15.54%** |
| F1 (Year) into 2 Bin | 14.16 % | **16.01%** | 13.19% | 12.04% |
| F1 (Year) into 10 Bin | 14.16 % | 12.82% | 12.50% | 10.38% |
| F1 to F20 into 10 Bins | 14.02 % | 13.24% | 14.48% | 13.05% |
| F1 to F20 into 4 Bins | 13.19 % | 13.38% | 14.81% | 13.01% |
| F1 to F20 into 2 Bins | 13.75 % | 14.30% | 14.21% | **15.45%** |
| F1 to F20 into 1 Bin | 12.59 % | 12.59% | 12.59% | **15.54%** |

**Table 1**: Prediction of weekdays on Brisbane data

**Result**: The following table (Table 2) shows the results of the prediction algorithms as they are applied across the four cities. Random Forest classifier performs best on Adelaide weather data, yielding **19.04 %** accuracy when the data is discretised into 10 bins.

| Unsupervised Discretisation of all Attributes (F1-F20) into 10 Bins for Brisbane, Adelaide, Perth and Hobart | | | | |
|---|---|---|---|---|
| Discretisation | Naïve Bayes Simple | Random Forests | J48 | IB1 |
| **Brisbane** | 14.02% | 13.24% | 14.48% | 13.05% |
| **Adelaide** | 15.42% | 17.95% | **19.04%** | 16.93% |
| **Perth** | 12.10% | 13.53% | 14.36% | 12.44% |
| **Hobart** | 13.81% | 11.89% | 13.39% | 8.75% |

**Table 2:** Prediction of weekdays across cities

2. Distinguishing between {Weekdays, Saturday, Sunday}

    **Result:** Both Naïve Bayes Simple and J48 classifiers were able to distinguish between Weekdays, Saturday and Sunday, with **73.34%** accuracy.

| Brisbane | | | | |
|---|---|---|---|---|
| Discretisation | Naïve Bayes Simple | Random Forests | J48 | IB1 |
| F1 to F20 into 4 Bins | **73.34%** | 69.14% | **73.34%** | 65.22% |

**Table 3:** Prediction for Weekdays, Saturday and Sunday on Brisbane data

3. Distinguishing between {Weekdays, Weekends}

    **Result:** Naïve Bayes Simple and Random Forests were the best performing algorithms both yielding 73.34% accuracy.

| Brisbane | | | | |
|---|---|---|---|---|
| Discretisation | Naïve Bayes Simple | Random Forests | J48 | IB1 |
| F1 to F20 into 4 Bins | **73.34%** | **73.34%** | 67.80% | 67.90% |

**Table 4:** Prediction for Weekdays and Weekends on Brisbane data

Please refer to **Table 4.1 - Result Set 1** provided in the Appendix, to view the complete set of results across the four cities.

**4.2 Will it rain tomorrow?**

In this section we try to predict weather conditions for a given day, i.e. we try to answer questions like *"Given tomorrow is Monday of the $2^{nd}$ week of December (for a given year), how likely*

*is it to rain?* In addition to rainfall we also try to predict the temperature and wind velocity for a given day.

In the following section we provide the results of our experiments.

**4.2 Result Set 2** – Predicting rain, average temperature and maximum wind for a given day using training and development data

In **Table 5** (below) we have used the prediction models to predict how likely it was to rain, what the average temperature and maximum wind velocity would be, on a given day in **Perth.**

| | | Perth | | | |
|---|---|---|---|---|---|
| Dist. | Class | NB | RF | J48 | IB1 |
| F1-F20 into 10 Bins | Rain | 73.39% | 84.46% | 84.41% | 80.37% |
| | Temp | 63.50% | 66.90% | 71.37% | 48.16% |
| | Wind | 20.02% | 19.58% | 20.07% | 15.99% |
| F1-F20 into 4 Bins | Rain | 76.68% | 87.06% | 87.41% | 81.65% |
| | Temp | 81.60% | 82.98% | 85.10% | 77.47% |
| | Wind | 41.96% | 40.29% | 43.04% | 35.91% |
| F1-F20 into 2 Bins | Rain | 87.90% | 89.03% | 89.72% | 86.72% |
| | Temp | 92.43% | 92.23% | **93.36%** | 92.13% |
| | Wind | 62.03% | 62.08% | 65.13% | 54.06% |

**Table 5:** Prediction for rainfall, wind and temperature using Perth data

By discretising the data into 10 bins we can not only say whether or not it will rain on a given day, **but we can also predict how much it will rain, if it does.** This provides results of higher resolution and hence adds more meaning to the results. The following table shows the different categories or ranges for the **Rainfall** attribute.

| Unsupervised Discretisation into 10 Bins | |
|---|---|
| **Nominal Label** | **Rainfall in mm** |
| **a = '(-inf-0.05]'** | 0 – 0.05 mm |
| **b = '(0.05-0.4]'** | 0.05 mm – 0.4 mm |
| **c = '(0.4-1.05]'** | 0.4 mm – 1.05mm |
| **d = '(1.05-2.85]'** | 1.05 mm – 2.85 mm |
| **e = '(2.85-5.05]'** | 2.85 mm – 5.05 mm |
| **f = '(5.05-8.15]'** | 5.05 mm – 8.15 mm |
| **g = '(8.15-10.85]'** | 8.15 mm – 10.85 mm |
| **h = '(10.85-17.05]'** | 10.85 mm – 17.05 mm |
| **i = '(17.05-30.85]'** | 17.05 mm – 30.85 mm |
| **j = '(30.85-inf)'** | 30.85 mm - higher |

Similarly, we can not only predict whether it would be warm on a given day, **but we can also predict how warm it is going to be or how windy it is going to be**. This additional resolution or "degree" adds much more meaning to results as compared to just answering "Yes" /"No" type questions.

**Observations:** In **Table 5** we can see that the accuracy of predictions **goes down** as the resolution of results goes up. **In other words, we can predict with higher accuracy between a smaller numbers of choices (coarse resolution).** But as we increase the number of choices **(higher resolution)** the accuracy goes down.

All of the four classifiers performed exceptionally well on Perth and Adelaide data for predicting the average temperature on a given day. With J48 classifiers yielding **93.36%** accuracy on Perth data discretized into 4 bins (Table 5).

Please refer to **Table 4.2 Result Set 2** provided in the Appendix, to view the complete set of results across the four cities.

5. **Conclusions**

The choice of discretisation technique(s) and classifier algorithm(s) used predominantly depends on the **context** and **type of available data**. Some algorithms are more suitable for nominal values while others perform best with numerical data.

It is hard to make a clear judgment based on the results obtained as part of this experiment, but in most cases **Random Forest**s and **J48** yielded in higher accuracy; slightly better results as compared to IB1 and noticeably better Naïve Bayes simple. Although in some instances Naïve Bayes yielded much higher accuracy, while the others were down.

From what we have observed in the test results, the accuracy of predictions **goes down** as the resolution of results goes up and vice versa. In other words, we can predict with higher accuracy between a smaller numbers of choices (coarse resolution). For instance, predicting between weekends and weekdays i.e. between 2 choices, resulted in much figures as compared to predicting the day of the week i.e. between 7 different choices.

One of the guiding principles is to ensure we provide as much meaning to our results as possible and to strike a balance between the resolution and accuracy of the results.

**References**

Zobel, J. and Dart, P. *Phonetic String Matching: Lessons from Information Retrieval*

Zhao, Y. and Zhang, Y. *Comparison of decision tree methods for finding active objects*

Khoo, A., Marom, Y., and Albrecht, D. *Experiments with Sentence Classification*

Witten, I. H. and Frank, E. *WEKA - Machine Learning Algorithms in Java,* Chapter 8, Data Mining: Practical Machine Learning Tools and Techniques with Java Implementations, 2000, Morgan Kaufmann Publishers

*Analytics: Decision Tree Induction*, Available at: http://gautam.lis.illinois.edu/monkmiddleware/public/monkmiddleware/public/index.html

*Naive Bayes*, Available at: http://docs.oracle.com/cd/B28359_01/datamine.111/b28129/algo_nb.htm

**4.1 Result Set 1** – Predicting the day of the week {Mon … Sun}, {Weekday, Sat, Sun} and {Weekday, Weekend} using training and development data

| Data Set | Properties | Class Attribute | Naïve Bayes Simple | Random Forest | J48 | IB1 |
|---|---|---|---|---|---|---|
| **Brisbane** | | | | | | |
| bris.dev.arff | Raw (No Pre-Processing) | Week day (F21) | 13.8838% | 13.2380% | 12.7768% | 12.1310% |
| bris.dev.prepro_14_0.arff | Replace Missing Values | Week day (F21) | 14.2989% | 13.3764% | 12.8229% | 12.1771% |
| bris.dev.prepro_13_0.arff | Supervised Discretisation of all attributes | Week day (F21) | 14.5295% | 14.5295% | 14.5295% | 15.5443% |
| bris.dev.prepro_14_1.arff | Unsupervised Discretisation of Year attribute into 2 Bins | Week day (F21) | 14.1605% | 16.0055% | 13.1919% | 12.0387% |
| bris.dev.prepro_14_2.arff | Unsupervised Discretisation of Year attribute into 10 Bins | Week day (F21) | 14.1605% | 12.8229% | 12.5000% | 10.3782% |
| bris.dev.prepro_14_3.arff | Unsupervised Discretisation of all Attributes (F1-F20) into 10 Bins | Week day (F21) | 14.0221% | 13.2380% | 14.4834% | 13.0535% |
| bris.dev.prepro_14_6.arff | Unsupervised Discretisation of all Attributes (F1-F20) into 4 Bins | Week day (F21) | 13.19% | 13.38% | 14.81% | 13.01% |
| bris.dev.prepro_14_4.arff | Unsupervised Discretisation of all Attributes (F1-F20) into 2 Bins | Week day (F21) | 13.7454% | 14.2989% | 14.2066% | 15.4520% |
| bris.dev.prepro_14_5.arff | Unsupervised Discretisation of all Attributes (F1-F20) into 1 Bin | Week day (F21) | 12.5923% | 12.5923% | 12.5923% | 15.5443% |
| bris.dev.prepro_14_7.arff | Unsupervised Discretisation of all Attributes (F1-F20) into 4 Bins | **Weekday/Sat/Sun** | **73.3395%** | 69.1421% | **73.3395%** | 65.2214% |
| bris.dev.prepro_14_8.arff | Unsupervised Discretisation of all Attributes (F1-F20) into 4 Bins | **Weekday/Weekend** | **73.3395%** | **73.3395%** | 67.8044% | 67.8967% |
| **Adelaide** | | | | | | |
| adel.dev.arff | Raw (No Pre-Processing) | Week day (F21) | 14.0361% | 17.8313% | 17.8313% | 10.8434% |
| adel.dev.prepro_1_0.arff | Replace Missing Values | Week day (F21) | 13.9759% | 18.6747% | 17.6506% | 14.6386% |
| adel.dev.prepro_13_0.arff | Supervised Discretisation of all attributes | Week day (F21) | 14.3976% | 14.3976% | 14.3976% | 13.6145% |
| adel.dev.prepro_1_1.arff | Unsupervised Discretisation of Year attribute into 2 Bins | Week day (F21) | 13.7349% | 18.7349% | 18.0723% | 15.4819% |
| adel.dev.prepro_1_2.arff | Unsupervised Discretisation of Year attribute into 10 Bins | Week day (F21) | 13.9759% | 17.7711% | 17.5301% | 14.5783% |
| adel.dev.prepro_1_3.arff | Unsupervised Discretisation of all Attributes (F1-F20) into 10 Bins | Week day (F21) | 15.4217% | 17.9518% | **19.0361%** | 16.9277% |
| adel.dev.prepro_1_6.arff | Unsupervised Discretisation of all Attributes (F1-F20) into 4 Bins | Week day (F21) | 14.8193% | **18.9759%** | 18.6145% | 18.4940% |
| adel.dev.prepro_1_4.arff | Unsupervised Discretisation of all Attributes (F1-F20) into 2 Bins | Week day (F21) | 13.8554% | 17.9518% | 15.9639% | 15.2410% |
| adel.dev.prepro_1_5.arff | Unsupervised Discretisation of all Attributes (F1-F20) into 1 Bin | Week day (F21) | 14.3976% | 14.3976% | 14.3976% | 13.6145% |
| adel.dev.prepro_1_7.arff | Unsupervised Discretisation of all Attributes (F1-F20) into 4 Bins | **Weekday/Sat/Sun** | 70.5422% | 69.6386% | 70.5422% | 64.8795% |
| adel.dev.prepro_1_8.arff | Unsupervised Discretisation of all Attributes (F1-F20) into 4 Bins | **Weekday/Weekend** | 70.5422% | 70.3614% | 70.5422% | 67.1084% |
| **Perth** | | | | | | |
| perth.dev.arff | Raw (No Pre-Processing) | Week day (F21) | 13.6252% | 13.0349% | 14.0187% | 14.5598% |
| perth.dev.prepro_1_0.arff | Replace Missing Values | Week day (F21) | 13.4776% | 12.1987% | 13.7236% | 9.9852% |
| perth.dev.prepro_13_0.arff | Supervised Discretisation of all attributes | Week day (F21) | 14.8549% | 14.8549% | 14.8549% | 14.4614% |
| perth.dev.prepro_1_1.arff | Unsupervised Discretisation of Year attribute into 2 Bins | Week day (F21) | 13.6252% | 12.5922% | 14.5106% | 11.9036% |
| perth.dev.prepro_1_2.arff | Unsupervised Discretisation of Year attribute into 10 Bins | Week day (F21) | 13.1825% | 12.1495% | 13.3792% | 9.4442% |
| perth.dev.prepro_1_3.arff | Unsupervised Discretisation of all Attributes (F1-F20) into 10 Bins | Week day (F21) | 12.1003% | 13.5268% | 14.3630% | 12.4447% |
| perth.dev.prepro_1_6.arff | Unsupervised Discretisation of all Attributes (F1-F20) into 4 Bins | Week day (F21) | 12.5922% | 13.7236% | 14.1663% | 14.3138% |
| perth.dev.prepro_1_4.arff | Unsupervised Discretisation of all Attributes (F1-F20) into 2 Bins | Week day (F21) | 13.1825% | 13.5268% | 14.1663% | 14.3630% |
| perth.dev.prepro_1_5.arff | Unsupervised Discretisation of all Attributes (F1-F20) into 1 Bin | Week day (F21) | 14.8549% | 14.8549% | 14.8549% | 14.4614% |
| perth.dev.prepro_1_7.arff | Unsupervised Discretisation of all Attributes (F1-F20) into 4 Bins | **Weekday/Sat/Sun** | 71.8151% | 60.4525% | 71.8151% | 64.6827% |
| perth.dev.prepro_1_8.arff | Unsupervised Discretisation of all Attributes (F1-F20) into 4 Bins | **Weekday/Weekend** | | 63.5514% | 71.8151% | 66.0108% |
| **Hobart** | | | | | | |
| hob.train.arff | Raw (No Pre-Processing) | Week day (F21) | 14.8408% | 13.7640% | 13.5768% | 12.4064% |
| hob.dev.prepro_1_0.arff | Replace Missing Values | Week day (F21) | 14.3258% | 13.6704% | 14.6067% | 10.7678% |
| hob.dev.prepro_13_0.arff | Supervised Discretisation of all attributes | Week day (F21) | 13.5300% | 13.5300% | 13.5300% | 15.5431% |
| hob.dev.prepro_1_1.arff | Unsupervised Discretisation of Year attribute into 2 Bins | Week day (F21) | 14.3727% | 15.2154% | 13.4363% | 12.3596% |
| hob.dev.prepro_1_2.arff | Unsupervised Discretisation of Year attribute into 10 Bins | Week day (F21) | 13.8109% | 11.8914% | 13.3895% | 8.7547% |
| hob.dev.prepro_1_3.arff | Nominal - All Attrb - 10 Bins | Week day (F21) | 13.5768% | 13.2959% | 13.8109% | 13.4363% |
| hob.dev.prepro_1_6.arff | Nominal - All Attrb - 4 Bins | Week day (F21) | 13.6236% | 13.4831% | 13.6704% | 13.2491% |
| hob.dev.prepro_1_4.arff | Nominal - All Attrb - 2 Bins | Week day (F21) | 13.4831% | 12.1255% | 13.4831% | 14.4663% |
| hob.dev.prepro_1_5.arff | Nominal - All Attrb - 1 Bins | Week day (F21) | 13.5300% | 13.5300% | 13.5300% | 15.5431% |
| hob.dev.prepro_1_7.arff | Unsupervised Discretisation of all Attributes (F1-F20) into 4 Bins | **Weekday/Sat/Sun** | 73.0805% | 68.2584% | 73.0805% | 65.7772% |
| hob.dev.prepro_1_8.arff | Unsupervised Discretisation of all Attributes (F1-F20) into 4 Bins | **Weekday/Weekend** | 73.0805% | 65.7303% | 73.0805% | 67.4625% |

**Table 4.1 – Result Set 1**

**4.2 Result Set 2** – Predicting rain, average temperature and maximum wind for a given day using training and development data

| Data Set | Properties | Class Attribute | Naïve Bayes Simple | Random Forest | J48 | IB1 |
|---|---|---|---|---|---|---|
| **Brisbane** | | | | | | |
| bris.dev.prepro_14_3.arff | Unsupervised Discretisation of all Attributes (F1-F20) into 10 Bins | Rainfall (F4) | 44.0498% | 65.4059% | 66.1439% | 55.3506% |
| | | Avg Temperature (F7) | 54.7509% | 61.5314% | 67.5738% | 48.2472% |
| | | Max Wind (F8) | 20.0185% | 20.2952% | 18.1734% | 17.2970% |
| bris.dev.prepro_14_6.arff | Unsupervised Discretisation of all Attributes (F1-F20) into 4 Bins | F4 - rain | 54.8432% | 67.5738% | 69.7878% | 58.7638% |
| | | Avg Temperature (F7) | 78.2749% | 82.1033% | 83.8561% | 75.2768% |
| | | Max Wind (F8) | 36.9926% | 37.9151% | 37.8690% | 34.7325% |
| bris.dev.prepro_14_4.arff | Unsupervised Discretisation of all Attributes (F1-F20) into 2 Bins | Rainfall (F4) | 66.5590% | 77.2140% | 73.8930% | 68.2657% |
| | | Avg Temperature (F7) | 91.2362% | 93.1734% | 93.3118% | 91.2362% |
| | | Max Wind (F8) | 58.0258% | 62.4539% | 65.5904% | 58.3948% |
| **Adelaide** | | | | | | |
| adel.dev.prepro_1_3.arff | Unsupervised Discretisation of all Attributes (F1-F20) into 10 Bins | Rainfall (F4) | 58.253% | 68.193% | 67.108% | 62.530% |
| | | Avg Temperature (F7) | 57.289% | 61.627% | 69.699% | 44.880% |
| | | Max Wind (F8) | 22.590% | 24.458% | 22.711% | 22.349% |
| adel.dev.prepro_1_6.arff | Unsupervised Discretisation of all Attributes (F1-F20) into 4 Bins | Rainfall (F4) | 65.422% | 70.542% | 70.482% | 63.735% |
| | | Avg Temperature (F7) | 81.566% | 82.831% | 83.072% | 74.337% |
| | | Max Wind (F8) | 40.602% | 44.458% | 39.819% | 40.482% |
| adel.dev.prepro_1_4.arff | Unsupervised Discretisation of all Attributes (F1-F20) into 2 Bins | Rainfall (F4) | 75.361% | 75.000% | 77.470% | 72.831% |
| | | Avg Temperature (F7) | **91.868%** | **90.843%** | **92.169%** | **91.205%** |
| | | Max Wind (F8) | 65.301% | 65.542% | 69.337% | 62.470% |
| **Perth** | | | | | | |
| perth.dev.prepro_1_3.arff | Unsupervised Discretisation of all Attributes (F1-F20) into 10 Bins | Rainfall (F4) | 73.39% | 84.46% | 84.41% | 80.37% |
| | | Avg Temperature (F7) | 63.50% | 66.90% | 71.37% | 48.16% |
| | | Max Wind (F8) | 20.02% | 19.58% | 20.07% | 15.99% |
| perth.dev.prepro_1_6.arff | Unsupervised Discretisation of all Attributes (F1-F20) into 4 Bins | Rainfall (F4) | 76.68% | 87.06% | 87.41% | 81.65% |
| | | Avg Temperature (F7) | 81.60% | 82.98% | 85.10% | 77.47% |
| | | Max Wind (F8) | 41.96% | 40.29% | 43.04% | 35.91% |
| perth.dev.prepro_1_4.arff | Unsupervised Discretisation of all Attributes (F1-F20) into 2 Bins | Rainfall (F4) | 87.90% | 89.03% | 89.72% | 86.72% |
| | | Avg Temperature (F7) | **92.43%** | **92.23%** | **93.36%** | **92.13%** |
| | | Max Wind (F8) | 62.03% | 62.08% | 65.13% | 54.06% |
| **Hobart** | | | | | | |
| hob.dev.prepro_1_3.arff | Unsupervised Discretisation of all Attributes (F1-F20) into 10 Bins | Rainfall (F4) | 64.4663% | 74.8127% | 73.8764% | 67.8839% |
| | | Avg Temperature (F7) | 60.9551% | 59.6910% | 67.4625% | 43.4457% |
| | | Max Wind (F8) | 18.9607% | 18.8202% | 17.2285% | 16.1985% |
| hob.dev.prepro_1_6.arff | Unsupervised Discretisation of all Attributes (F1-F20) into 4 Bins | Rainfall (F4) | 66.8071% | 77.4813% | 77.1067% | 72.0506% |
| | | Avg Temperature (F7) | 79.4944% | 81.3202% | 83.0524% | 72.0037% |
| | | Max Wind (F8) | 35.2996% | 37.5000% | 40.8240% | 33.4270% |
| hob.dev.prepro_1_4.arff | Unsupervised Discretisation of all Attributes (F1-F20) into 2 Bins | Rainfall (F4) | 76.9195% | 81.1798% | 83.5206% | 72.7996% |
| | | Avg Temperature (F7) | 91.7603% | 92.0880% | 92.5562% | 90.6367% |
| | | Max Wind (F8) | 60.6273% | 61.8446% | 65.4494% | 58.0056% |

Table 4.2 – Result Set 2